\documentclass{article}
\usepackage{spconf,amsmath,graphicx}
\newcommand{\ie}{\textit{i}.\textit{e}.}
\newcommand{\eg}{\textit{e}.\textit{g}.}
\usepackage{amssymb}

\title{Towards Multi-domain Face Landmark Detection with Synthetic data from Diffusion model}

\name{Yuanming Li, Gwantae Kim, Jeong-gi Kwak, Bon-hwa Ku, Hanseok Ko\sthanks{* The correspongding author.} \thanks{This work was supported by the "Development of cognitive/response advancement technology for AI avatar commercialization" project funded by the Brand Engagement Network (BEN) [Q2312881].}}
\address{School of Electrical Engineering, Korea University, Seoul, Korea}
\begin{document}

\maketitle
\begin{abstract}
Recently, deep learning-based facial landmark detection for in-the-wild faces has achieved significant improvement. However, there are still challenges in face landmark detection in other domains (\eg{} cartoon, caricature, etc). This is due to the scarcity of extensively annotated training data. To tackle this concern, we design a two-stage training approach that effectively leverages limited datasets and the pre-trained diffusion model to obtain aligned pairs of landmarks and face in multiple domains. In the first stage, we train a landmark-conditioned face generation model on a large dataset of real faces. In the second stage, we fine-tune the above model on a small dataset of image-landmark pairs with text prompts for controlling the domain. Our new designs enable our method to generate high-quality synthetic paired datasets from multiple domains while preserving the alignment between landmarks and facial features. Finally, we fine-tuned a pre-trained face landmark detection model on the synthetic dataset to achieve multi-domain face landmark detection. Our qualitative and quantitative results demonstrate that our method outperforms existing methods on multi-domain face landmark detection.

\end{abstract}

\section{Introduction}
\label{sec:introduction}

Facial landmark detection is a critical area in computer vision, with applications \cite{yi2022animating, deng2019accurate, tautkute2018know} in a wide range of technologies, including 3D facial reconstruction, face recognition, expression recognition, and augmented reality (AR)/virtual reality (VR). While supervised landmark detection has made significant progress for in-the-wild faces \cite{zhang2014facial, bulat2017far, dong2018style, feng2018wing, huang2021adnet, zhou2023star}, facial landmark recognition for new image domains (such as art, cartoon, and caricature) still faces challenges. The main challenge is the lack of paired image-landmark datasets from other domains. Data collection is also a time-consuming and expensive process.

Existing methods mainly use geometry warping and style translation to transform existing face landmark datasets into a data distribution that is similar to the target domain. Yaniv \textit{et al.} \cite{yaniv2019face} propose an approach for augmenting natural face images artistically, enabling the training of networks to detect landmarks in artistic portraits. Facial landmark datasets \cite{sagonas2016300} are transformed into "artistic face" data using a technique known as geometric-aware style transfer. This process combines a gram matrix-based method \cite{gatys2015neural} for transferring artistic styles with geometric warping to achieve the transfer of facial features and geometry. Recently, Sindel \textit{et al.} \cite{sindel2022artfacepoints} introduced an approach to detect facial landmarks in high-resolution paintings and prints. Similar to the previous method, they also use geometric-aware style transfer to perform augmentation on the dataset. What sets it apart is that it employed CycleGAN \cite{zhu2017unpaired, kwak2021adverse, li2021adaptive} to perform style transfer on the images. However, when dealing with datasets that exhibit a substantial domain gap from real images, such as cartoons, caricatures, etc., the geometric-aware style transfer used in these methods tends not to work well, leading to suboptimal results in terms of image translation. Cai \textit{et al.} \cite{cai2021landmark} also proposed an automatic method for 3D caricature reconstruction. The method is to first regress a 3D face model and orientation and then project 3D landmarks and orientation to recover 2D landmarks. However, their method is only applicable to caricatures and requires a large amount of data to train.

Recent advances in text-to-image synthesis \cite{rombach2022high, podell2023sdxl} have shown impressive performance and creativity, enabling the conversion of text descriptions into visually appealing creations. These generative systems excel in a variety of areas, including rendering realistic and diverse appearances, powerful editing, etc. In more recent developments, ControlNet \cite{zhang2023adding} extends the capabilities of pretrained diffusion models using auxiliary input conditions. ControlNet proficiently captures task-specific conditions like pose, depth map, and semantic map, demonstrating robustness even when trained on limited datasets. 


In this paper, we tackle multi-domain face landmark detection by leveraging synthetic data generated through a text-to-image diffusion model. Firstly, our goal is to generate high-quality data pairs of multi-domain face images and their landmarks. We propose a two-stage training approach that generates high-quality data pairs of multi-domain face images and their landmarks using a small dataset and pre-trained text-image model based on ControlNet. In the first stage, we train ControlNet on a large dataset of real-world faces \cite{karras2019style}. We use facial landmarks as a condition aligned with the face image as a condition for generating the face image. In the second stage, the pre-trained model is fine-tuned using a small multi-domain face dataset to generate face images of diverse domains. Secondly, we created a large multi-domain facial landmark dataset of 25 styles, each of which contains 400 images with annotations. We control the geometric characteristics and styles of face images by editing landmarks and text prompts. Finally, we fine-tune the existing face landmark detection model \cite{zhou2023star} on this dataset, which achieved state-of-the-art performance on the ArtFace\cite{yaniv2019face} and Caricature datasets\cite{cai2021landmark}.

\section{Proposed Method}
Our method is based on the pre-trained text-to-image model, \ie{}, latent diffusion model \cite{rombach2022high}. In this section, we first give a brief introduction to the details of the latent diffusion model in Sec. \ref{ldm}. Next, we present an overview of our landmark-guided diffusion model in Sec. \ref{landmarkldm} and describe our procedure for generating synthetic datasets in Sec. \ref{data}. Finally, we describe the fine-tuning process for the face landmark detector in Sec. \ref{detection}.

\begin{figure*}[!t]
\begin{center}
\centering\includegraphics[scale=0.43]{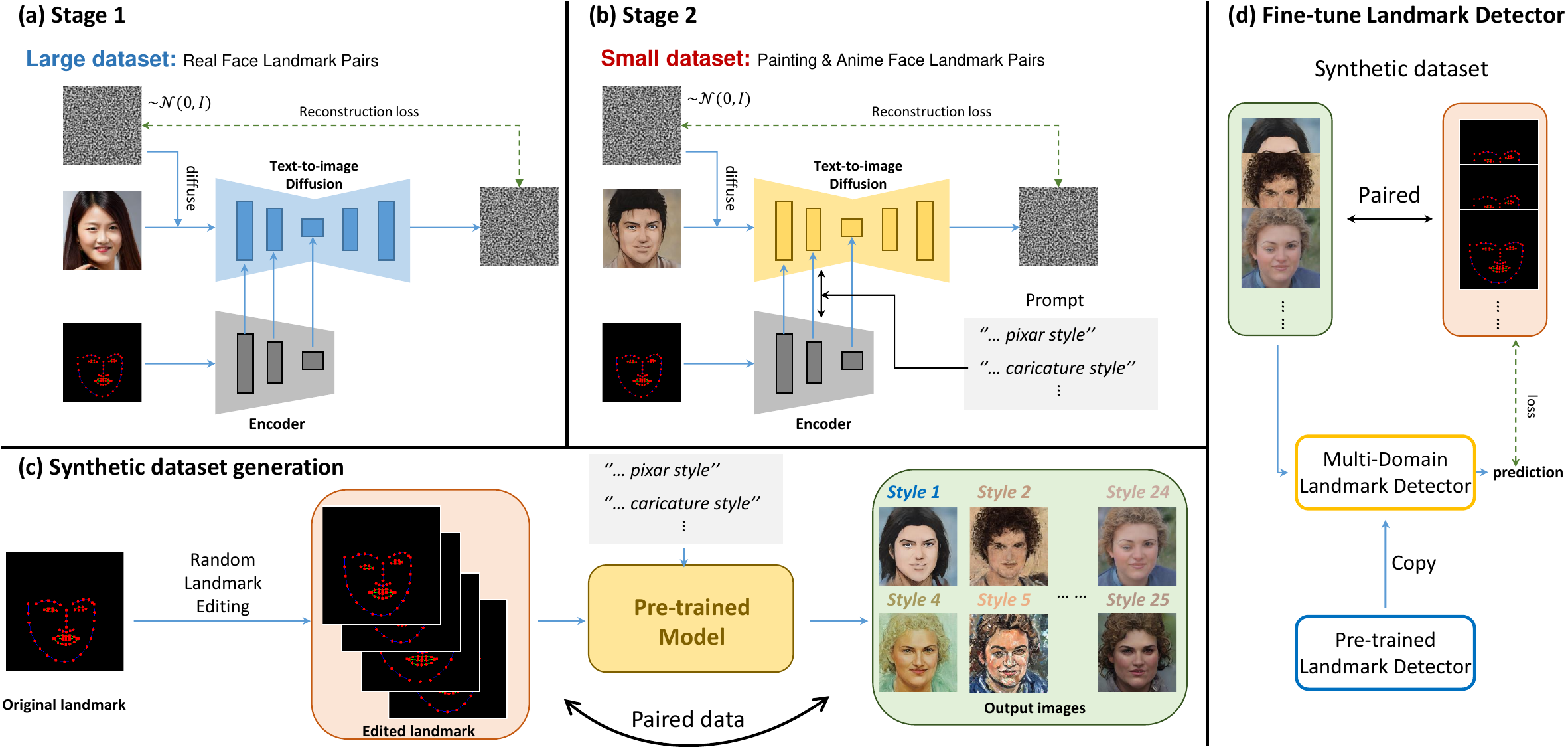}
\end{center}
\setlength{\abovecaptionskip}{0pt}
\vspace{-0.6cm}
   \caption{Overview of our proposed framework. We propose a two-stage framework for creating a high-quality multi-domain face from a face landmark with diffusion prior. (a) First stage, we train our model with a pair of real faces and landmarks. (b) Second stage, we further fine-tune the model with a small multi-domain dataset. (c) We generate diverse facial data by controlling both the text and landmarks. (d) We fine-tune the pre-trained landmark detector with the synthetic dataset.}

\label{arch}
\end{figure*}

\subsection{Latent Diffusion Model}\label{ldm}
The Latent Diffusion Model (LDM) is a diffusion model that has recently exhibited exceptional performance in the realm of image synthesis. It comprises two main components: an autoencoder and a diffusion model. The autoencoder, consisting of an encoder $E$ and a decoder $D$, learns to reconstruct images. The encoder first projects an image x into a lower-dimensional latent space: $z = E(I)$, and the decoder reconstructs the original image from the latent space: $\hat{I} = D(z)$. Then, the Denoising Diffusion Probabilistic Model (DDPM) \cite{ho2020denoising} is employed to generate the latent $z$. DDPM adds noise to $z_0$ gradually with $T$ timesteps to mimic diffusion in non-equilibrium thermodynamics, which can be modeled as:
\begin{equation}
    q(z_t|z_{t-1}) := N(z_t;\sqrt{1-\beta_{t}}z_{t-1},\beta_{t})
\end{equation}
The Markov chain is a gradually forward noising process by a predefined noise schedule ${\beta_1,...,\beta_T}$. Subsequently, they proceed to construct and train a U-Net model to acquire an understanding of the reverse diffusion process, denoted as:
\begin{equation}
    q(z_t|z_0) = N(z_t;\sqrt{\bar{\alpha}_{t}}z_0,(1-\bar\alpha_{t})I)
\end{equation}

The reverse process starts from $z_T$ , which is close to an isotropic Gaussian. The U-Net then is trained to predict $z_{t-1}$ from $z_t$ by estimating the true posterior $q(z_{t-1}|z_t)$ which can be modeled as:
\begin{equation}
    p_\theta(z_{t-1}|z_t) = N(z_{t-1};\mu_{\theta}(z_t, t), \sigma_{\theta}(z_t, t))
\end{equation}
At each time step, a random noise $\epsilon$ is drawn from a diagonal Gaussian distribution, and a time-conditioned denoising model $\theta$ is trained to predict the added noise in each timestep $t$ with a simple MSE error:
\begin{equation}
   L(\theta) := ||\epsilon_{\theta}(z_t, t)-\epsilon||^2_2
\end{equation}

\subsection{Landmark-guided Latent Diffusion Model} \label{landmarkldm}
Given a landmark $L$ of dimension $N\times2$, where $N$ is the number of landmarks, and a prompt $\boldsymbol{c}$ description (\eg{}, “\textit{cartoon style, pixar style}”, in Fig. \ref{arch}). Our goal is to generate a face that maintains the same facial structure by inputting exaggerated landmarks and style-relevant prompts. Due to the lack of a large-scale multi-domain facial landmark annotation dataset, we propose a two-stage training framework suitable for small datasets. The details of the small dataset will be discussed in Sec. \ref{small}.

As shown in Fig. \ref{arch} (a) and (b), our method contains two stages of training for different purposes. \textbf{\textit{First stage}}. We prepare a training dataset $D^{src}={z^{src}, L^{src}}$ with $256^2$ Flickr-Faces-HQ (FFHQ) for first-stage training, which consists of the diffusion latent $z^{src}=E(x^{src})$ from real face $x^{src}$ encoded by VQGAN encoder \cite{esser2021taming}, and its landmark $L^{src}$. Note that we directly use visualized landmark images with coordinates as training conditions. To facilitate better recognition of semantic contents from landmarks for generated images, we add empty strings for training at this stage. \textbf{\textit{Second stage}}. We create a small scale multi-domain dataset $D^{trg}={z^{trg}, L^{trg}, c^{trg}}$, which consists of the diffusion latent $z^{trg}$ from multi-domain face $x^{trg}$, the landmark $L^{trg}$ of $x^{trg}$, and text-prompt $c^{trg}$. In addition to the landmark and image data pairs, we added text prompts, such as “\textit{a portrait of cartoon style},” to control the style of the output. After fine-tuning, we can generate exaggerated multi-domain faces by editing the coordinates of landmarks, while keeping the landmarks and faces aligned. We use the same training objective \cite{zhang2023adding} in both stages.

\subsection{Synthetic Dataset Geneartion}\label{data}

As shown in Fig. \ref{arch} (c), we random sample the landmarks from the training dataset and edit them according to the following guidelines. We selected the following editing attributes: chubby, linear transform, open eyes, raised eyebrow, stretched nostrils, and eye/nose/mouth shifting. When editing landmarks, we random sample two of them each time. Then, we control the domain through text prompts, and we have a total of 25 different styles to choose from. We generated 400 images for each style category, thus, our synthetic dataset contains a total of 10,000 images. The generated samples are shown in Fig. \ref{samples}. Compared to some classic GAN methods, they often encounter difficulties in generating multi-domain images. Therefore, a new model needs to be trained for each new domain. In this regard, we only need to train one model to control the generation of a wide variety of images.

\subsection{Fine-tuning Landmark Detection Model}\label{detection}
In contrast to training a new model, we found that fine-tuning an existing model with synthetic data produces better results. Therefore, we employ a model \cite{zhou2023star} with high performance for real-world facial landmark detection as a pre-trained model. The landmark regression method is based on a stacked Hourglasses Networks (HGs) \cite{zhang2014facial} that generates $N$ heatmaps, each of which is a probability distribution over the predicted facial landmarks. Finally, the landmark locations are fine-tuned based on STAR loss \cite{zhou2023star}.

\section{Experimental}
\textbf{\textit{Diffusion model}}. We implement our method based on the official codebase of Stable Diffusion \cite{rombach2022high} and the publicly available 1.4 billion parameter T2I model. We first train our model for 200k steps on the FFHQ dataset\cite{karras2019style}. Then, we train for 100k steps on our small multi-domain dataset. The resolution of 256 $\times$ 256 from the input landmark. The batch size and learning rate are set to 4 and $10^{-5}$. The training process is performed on only one NVIDIA RTX Titan GPU and can be completed within one day. At inference, we apply DDIM sampler \cite{song2020denoising} with classifier-free guidance \cite{ho2022classifier} in our experiments for landmark-guided face generation. \textbf{\textit{Face landmark detector}}. Learning rate and batch size are set to $10^{-4}$ and 32. We use ADAM optimization \cite{kingma2014adam} with $\beta_1 = 0.9$, $\beta_2 = 0.999$.

\noindent\textbf{Baseline}. We make comparisons to foa\cite{yaniv2019face}, ArtFacePoint\cite{rombach2022high}, STAR\cite{zhou2023star}, and CariFace\cite{cai2021landmark}. All the models and their pre-trained models are provided on their official Github website. Since CariFace is primarily designed for landmark detection in caricature-style images, it largely failed when tested on the ArtFace dataset. Therefore, we did not showcase its results on ArtFace.
\vspace{-0.3cm}

\subsection{Dataset}
\vspace{-0.3cm}


\noindent\textbf{FFHQ}: The Flickr-Faces-HQ Dataset, notable in high-fidelity GAN projects like StyleGAN \cite{karras2019style}, contains $70,000$ high-resolution ($1024\times1024$) images, resized to $256\times256$ for our training purposes. \noindent\textbf{ArtFace}: The ArtFace dataset \cite{yaniv2019face}, a collection of art paintings, comprises 160 images ($256\times256$) with 68 landmarks, each representing various artistic styles. \noindent\textbf{CariFace}: This dataset features landmark annotations for caricatures \cite{cai2021landmark}, with 6,240 images for training and 1,560 for testing. \noindent\textbf{Small multi-domain dataset}\label{small}: We created a small dataset ($<$1k) of images using other generative model \cite{yang2022Pastiche} and a subset of the Artface dataset to train our diffusion model. We randomly selected 112 (70\%) images from the ArtFace dataset and 100 (1.6\%) images from the CariFace dataset for the training set. In addition, we used a pre-trained DualstyleGAN to generate $50-100$ synthetic images for each of the 8 styles. The total number is 984. We then used \cite{yaniv2019face} to annotate these images and manually refined the coordinates. 

\begin{figure}[!t]
\begin{center}
\centering\includegraphics[scale=0.4]{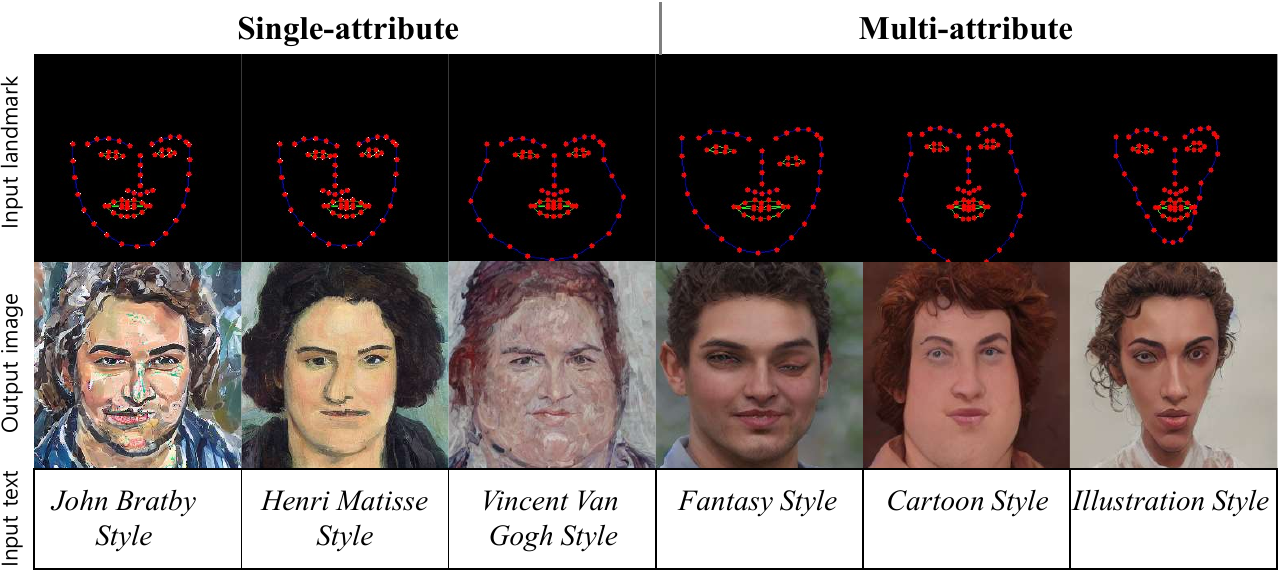}
\end{center}
\setlength{\abovecaptionskip}{0pt}
\vspace{-0.8cm}
   \caption{Sample results of our method.}
\label{samples}
\end{figure}

\section{Results and Discussion}
\begin{figure}[!t]
\begin{center}
\centering\includegraphics[scale=0.4]{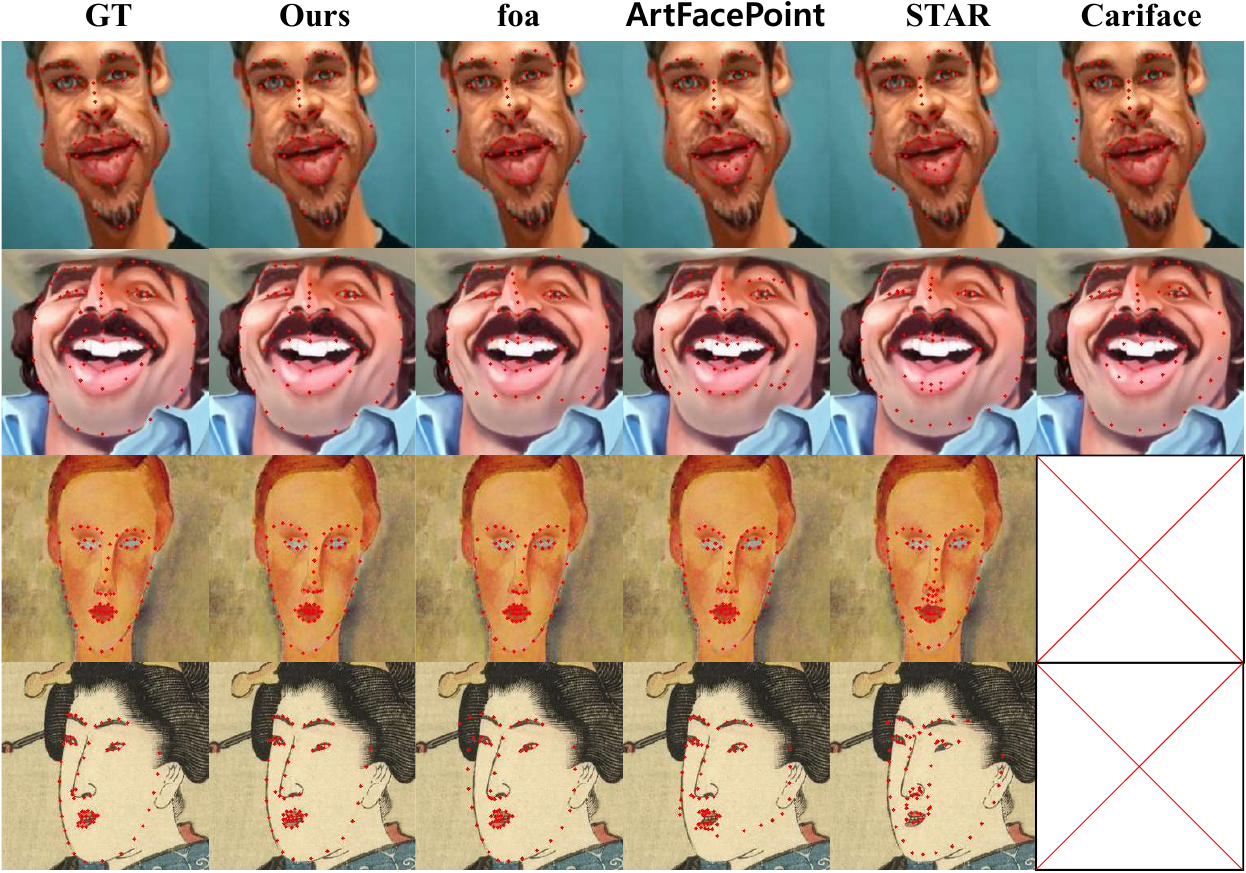}
\end{center}
\setlength{\abovecaptionskip}{0pt}
\vspace{-0.8cm}
   \caption{Qualitative comparison between evaluated methods.}
\label{comparision}
\end{figure}

\vspace{-0.3cm}
\noindent\textbf{Qualitative results}. In Fig. \ref{samples}, we show some images generated from the input of edited landmarks and text. As can be seen from the results, our generated samples are well aligned with the landmarks after the landmarks are exaggeratedly edited. Qualitative results can be visualized in Fig. \ref{comparision}. For caricature results, foa and ArtFacePoint exhibited large errors in the prediction of facial contours. Although the Cariface model performs well compared to other methods, our method is still more accurate in capturing mouth details and facial contours. ArtFacePoint and foa both perform well on simple painting images. However, our method is significantly better at capturing facial details in images with a large domain gap. STAR is trained on real-world face images, which is not a fair comparison here. However, we do this to demonstrate the effectiveness of our synthetic dataset.

\noindent\textbf{Quantitative results}. We report relative Normalized Mean Error (NME), Failure Rate (FR), and Area Under Curve (AUC) similar to \cite{zhou2023star} for Artface and CariFace dataset experiments in Table \ref{nme}. Notably, we achieved state-of-the-art results on the ArtFace dataset. It is important to clarify that our model's training was conducted using a significantly smaller dataset, consisting of only 100 images for the diffusion model training, compared to over 6,000 images utilized by CariFace. This aspect underscores the efficiency of our approach, as it demonstrates our method's capability to achieve competitive, and in some cases superior, results with substantially less training data. The slight underperformance on the CariFace dataset, compared to the CariFace method, can be attributed to the limited size of our training set, yet the results remain robust and indicative of our method's efficacy in multi-domain facial landmark detection.

\begin{figure}[!t]
\begin{center}
\centering\includegraphics[scale=0.36]{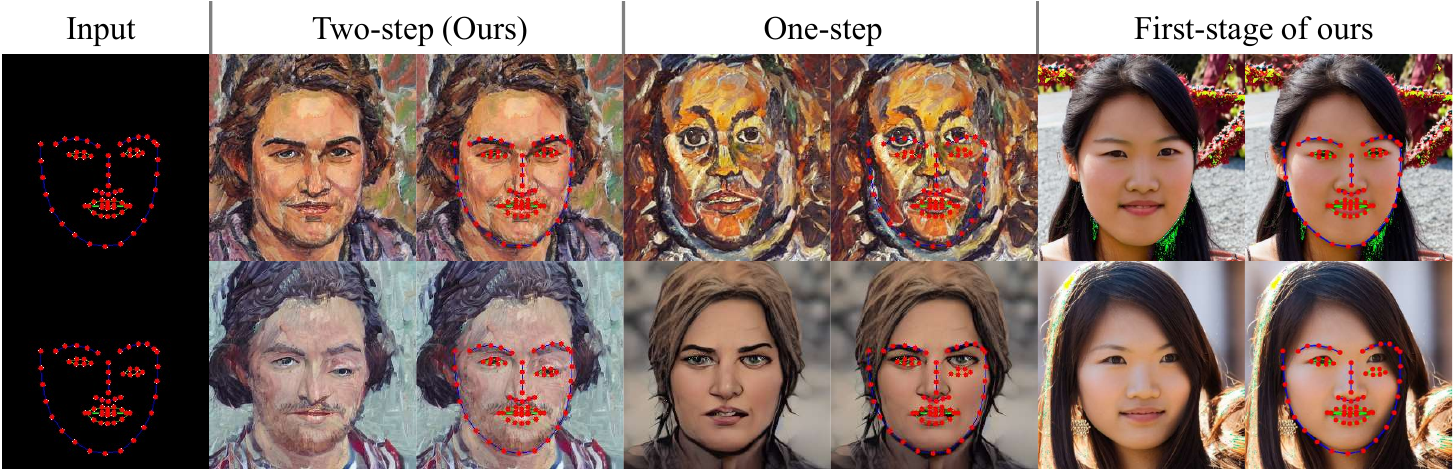}
\end{center}
\setlength{\abovecaptionskip}{0pt}
\vspace{-0.8cm}
   \caption{Ablation study of our method.}
\label{ablation}
\end{figure}

\noindent\textbf{Ablation results}. As shown in Fig. \ref{ablation}, we present the results of the ablation study. When we directly train a diffusion model on a small-scale dataset (One-step), the generated faces still retain the style of the input prompt. However, the generated faces do not match the input landmarks semantically. The landmarks fail to exert control over the generated images due to their lack of alignment with the generated faces. Training on a large-scale dataset of real-world face images aligns faces with landmarks effectively. Nevertheless, introducing exaggerated landmarks as input deviates from the spatial distribution seen in our familiar face data, resulting in some inherent bias. With our two-stage approach, we not only have the ability to generate multi-domain faces using text prompts but can also produce aligned face images even when given exaggerated landmarks as input.

\begin{table}[htbp]
 \setlength{\tabcolsep}{0.5mm}
  \centering
  \caption{Quantitative comparison with evaluated baselines.}
  \begin{tabular}{*{7}{c}}
      \hline
      \multicolumn{1}{c}{}& \multicolumn{3}{c}{ArtFace} &\multicolumn{3}{c}{CariFace} \\
      \hline
      Metric & $NME$  & $FR_{10\%}$  & $AUC_{10\%}$  & $NME$    & $FR_{10\%}$   & $AUC_{10\%}$  \\
      \hline
      Ours   & 4.64 & 2.26 & 0.5548 & 5.54 & 6.29 & 0.4838\\
      foa    & 4.69 & 3.75 & 0.5388 & 8.26 & 22.31 & 0.2997\\
      ArtFace& 6.50 & 10.62 & 0.4573 & 12.04 & 44.41 & 0.1476\\            
      CariFace& -     &      - &      - & 4.54 & 0.71 & 0.5477\\
      STAR   & 6.20 & 13.21 & 0.5142 & 7.16 & 13.73 & 0.3875\\
      \hline
  \end{tabular}
\label{nme}
\end{table}

\vspace{-0.5cm}
\section{Conclusions}
\vspace{-0.3cm}
We introduced a novel two-stage approach for creating synthetic datasets for multi-domain facial landmark detection, effectively utilizing limited datasets and pre-trained diffusion models. Our method generates high-quality, domain-spanning synthetic datasets, maintaining alignment between landmarks and facial features. We then fine-tuned a pre-trained landmark detection model on this dataset. Experimental results show our method outperforms others in accuracy on both ArtFace and CariFace datasets.

\small
\bibliographystyle{IEEEbib}
\bibliography{strings,refs}

\end{document}